\documentclass{article}

\usepackage{iclr2024_conference,times}
\iclrfinalcopy

\usepackage{comment}

\usepackage{graphicx} 
\usepackage{hyperref}
\usepackage{multirow}
\usepackage{subfig}
\hypersetup{
    colorlinks=true,
    linkcolor=blue,
    filecolor=magenta,      
    urlcolor=cyan,
    pdftitle={compbio},
    pdfpagemode=FullScreen,
    }
\usepackage{listings}
\usepackage{soul}
\usepackage{xcolor}
\usepackage{amsmath}
\usepackage{tikz}
\usepackage{mathdots}
\usepackage{yhmath}
\usepackage{cancel}
\usepackage{color}
\usepackage{siunitx}
\usepackage{array}
\usepackage{multirow}
\usepackage{amssymb}
\usepackage{tabularx}
\usepackage{extarrows}
\usepackage{booktabs}
\usepackage{url}
\usepackage{adjustbox}
\usepackage{makecell}
\usepackage{tcolorbox}

\DeclareMathAlphabet{\mathpzc}{OT1}{pzc}{m}{it}

\newcommand{\FT}{\mathpzc{F}\hspace{0.005pt}\mathpzc{T}}
\newcommand{\LP}{\mathpzc{L}\hspace{1pt}\mathpzc{P}}

\definecolor{codegreen}{rgb}{0,0.6,0}
\definecolor{codegray}{rgb}{0.5,0.5,0.5}
\definecolor{codepurple}{rgb}{0.58,0,0.82}
\definecolor{backcolour}{rgb}{0.95,0.95,0.92}

\lstdefinestyle{mystyle}{
  backgroundcolor=\color{backcolour},   commentstyle=\color{codegreen},
  keywordstyle=\color{magenta},
  numberstyle=\tiny\color{codegray},
  stringstyle=\color{codepurple},
  basicstyle=\ttfamily\footnotesize,
  breakatwhitespace=false,         
  breaklines=true,                 
  captionpos=b,                    
  keepspaces=true,                 
  numbers=left,                    
  numbersep=5pt,                  
  showspaces=false,                
  showstringspaces=false,
  showtabs=false,                  
  tabsize=2
}

\usepackage{booktabs}       

\title{Foundation Model's Embedded Representations May Detect Distribution Shift}

\author{%
\textbf{Max Vargas}$^{1,\footnote{Equal Contributions}}$ \quad
  \textbf{Adam Tsou}$^{1,2,*}$ \quad
   \textbf{Andrew Engel}$^{1}$ \quad \textbf{Tony Chiang}$^{1,3,4}$\\
   $^1$Pacific Northwest National Laboratory\quad
  $^2$Stony Brook University \\ \quad $^3$University of Washington \quad $^4$University of Texas, El Paso\\
  \texttt{\{max.vargas,andrew.engel,tony.chiang\}@pnnl.gov}\\
  \texttt{\{adam.tsou\}@stonybrook.edu}\\
  \texttt{* Equal Contribution}
}

\date{January 2024}

\begin{document}

\maketitle

\begin{abstract}
    Sampling biases can cause distribution shifts between train and test datasets for supervised learning tasks, obscuring our ability to understand the generalization capacity of a model. This is especially important considering the wide adoption of pre-trained foundational neural networks --- whose behavior remains poorly understood --- for transfer learning (TL) tasks. We present a case study for TL on the Sentiment140 dataset and show that many pre-trained foundation models encode different representations of Sentiment140's manually curated test set $M$ from the automatically labeled training set $P$, confirming that a distribution shift has occurred. We argue training on $P$ and measuring performance on $M$ is a biased measure of generalization. Experiments on pre-trained GPT-2 show that the features learnable from $P$ do not improve (and in fact hamper) performance on $M$. Linear probes on pre-trained GPT-2's representations are robust and may even outperform overall fine-tuning, implying a fundamental importance for discerning distribution shift in train/test splits for model interpretation. 
\end{abstract}


\begin{section}{Introduction}


Foundation models \cite{brownGPT3, llama2, liu2023improvedllava, rombach2021highresolution} have quickly integrated themselves into the standard machine learning development stack, in particular for their adaptability to specialized tasks \cite{image-finetuning, fewshot-learning}. Under the umbrella of transfer learning (TL), this specialization is often performed through fine-tuning after pre-training on a diverse corpus \citep{refinedweb}, 
which is believed to result in highly generalizable feature representations \citet{liu2023improvedllava}. While a variety of open-source foundation models have been released over the last decade, its unknown what their representations encode or do not encode from the original high-dimensional and feature-rich natural language input. This question is especially important under distribution shift, where the train and test datasets come from different populations (e.g., due to a biased sampling or labeling procedure).

The broad, general knowledge encoded in foundation models can provide new insight in dealing with sampling bias effects, a persistent problem in statistical science known to be present in many datasets \cite{10.5555/1462129}. This article addresses these concerns involving distribution shift in the context of transfer learning using the Sentiment140 dataset \citet{Go2009TwitterSC}, a sentiment classification dataset whose sampling procedures differs between train/test splits. 
Our observations emphasize the importance of sampling train and test data from the same population, illustrating a point of caution when fine-tuning on data which is only semantically similar to the testing task:
the presence of distribution shifts makes it unclear whether fine-tuning will at all boost the performance of a foundation model. To be explicit, the main contributions of our article are:
\begin{itemize}
\item We show a simple examination of the final feature embedding using principal component analysis (PCA) is able to detect the train and test data are sampled from different populations (see Figure~\ref{PLOT1}), giving us a data-centric method to probe distribution shifts.
\item We proceed to fine-tune foundation models to study the effect of this distribution shift. Our experiments show worse generalization when fine-tuned on a distribution assumedly closer to our test distribution compared to the original pre-training corpora. In other words: fine tuning under distribution shift obscured the capacity for the model to generalize on the test population and confounds this capacity for model robustness under the distribution shift. This reiterates the necessity to sample from the same population inference is performed on.
\end{itemize}

 \begin{figure}[!ht]
    \centering
        
        \includegraphics[width=0.3\linewidth]{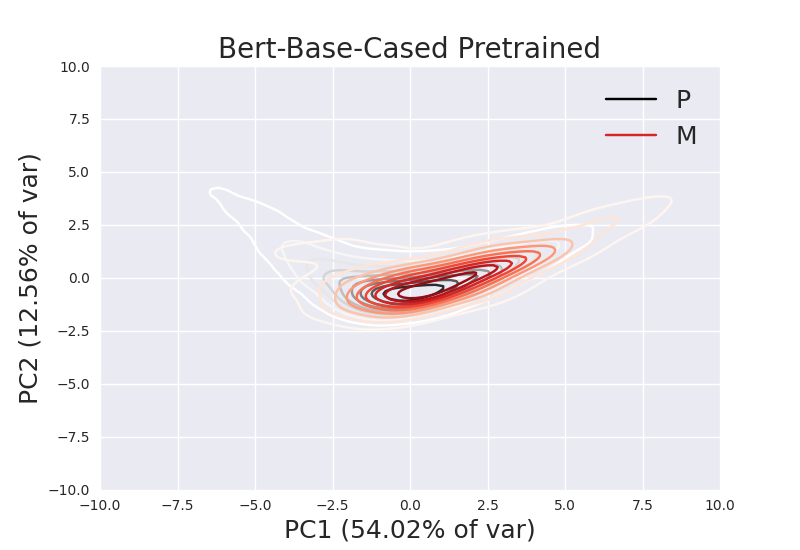}
        \includegraphics[width=0.3\linewidth]{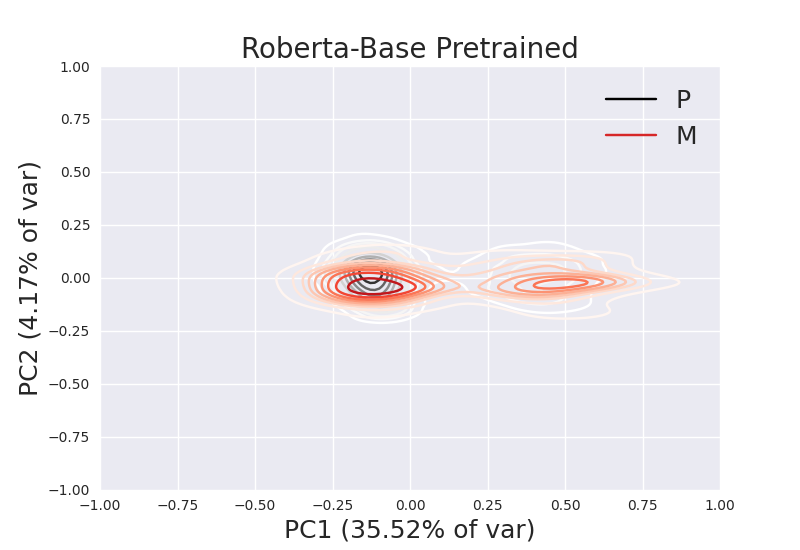}
        \includegraphics[width=0.3\linewidth]{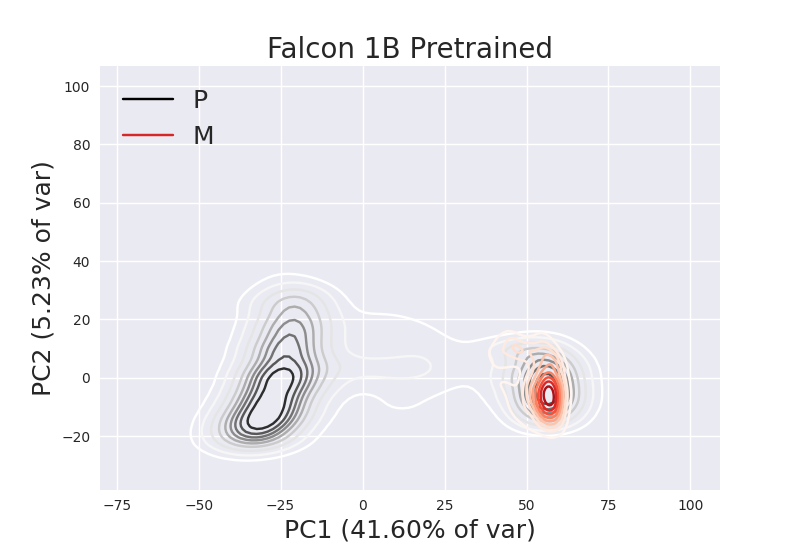}
        \\
        \includegraphics[width=.3\linewidth]{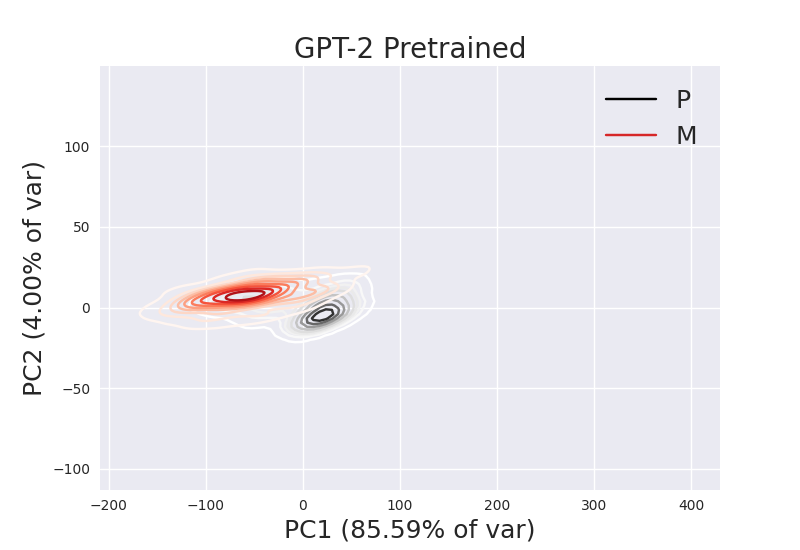}
        \includegraphics[width=0.3\linewidth]{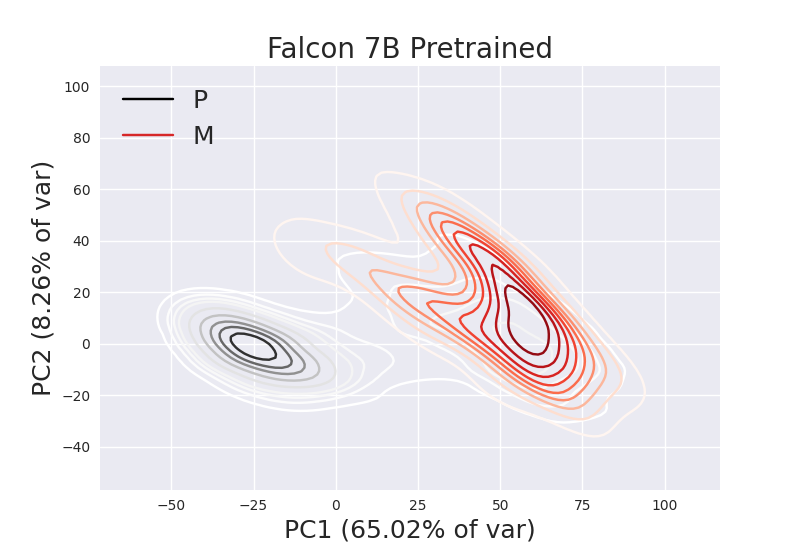}
        \includegraphics[width=0.3\linewidth]{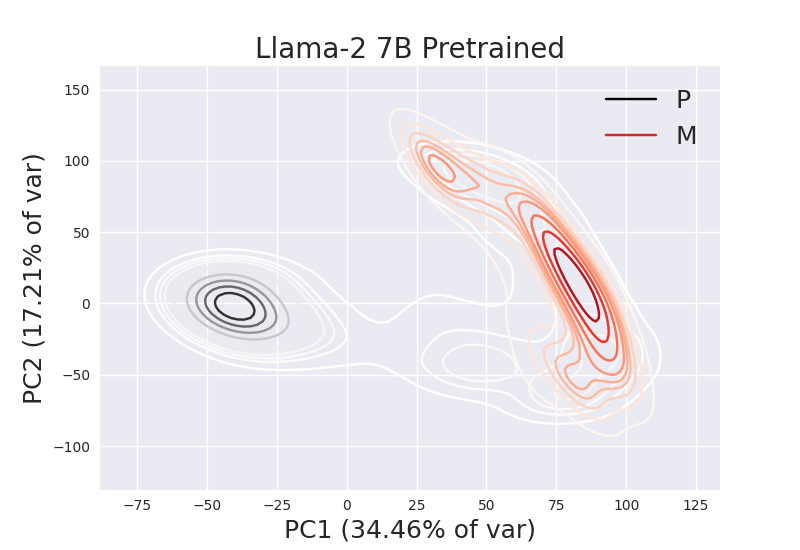}
    \caption{Kernel Density Estimates of the two largest principal components of the pre-trained final embedding representation of the automatically labeled training dataset \(P\) and the manually curated testing dataset \(M\) using various LLMs. Sub-figures are ordered in number of increasing model parameters. 
    }\label{PLOT1}
  \end{figure}
  
\end{section}
\section{Related Work}


\textbf{Transfer learning.} Adapting trained models to new tasks is a popular technique, believed to improve performance by leveraging knowledge gained from a related domain or otherwise saving costs compared to training from scratch \citet{limitsoftransfer}. Two common approaches are full fine-tuning and linear probing. In the former, all layers are allowed to update \citet{Yosinski2014HowTA}, while the ladder only allows a linear, task-specific layer to update. Linear probing has been studied for its explanatory uses \citet{Belinkov2021ProbingCP}, \citet{Alain2016UnderstandingIL}, \citet{Chen2020GenerativePF} and to provide relative baselines \citet{Kumar2022FineTuningCD}. Our experiments compare linear probes and full fine-tuning in order to measure their relative effectiveness on shifted train and test data; see \citet{Evci2022Head2ToeUI} for a different comparison of linear probes and fine-tuning.


\textbf{Foundation Models} Commonly built on the transformer architecture for vision and language contexts, these models first go though pre-training on a large, general use data corpus before being specialized to a downstream task \citet{transformer,brownGPT3,rombach2021highresolution}. Though the pre-training corpus is typically sourced from the open web, many details are often closed the public \citet{geminiteam2023gemini,llama2}. The methods provided here implicitly take advantage of the knowledge gained from these training corpora to test for distribution shifts.

\textbf{Representation Learning.} Also known as feature learning, this focuses on learning useful features directly from raw data instead of hand-crafting features for a specific task \citet{10.1109/TPAMI.2013.50}. The problem of explaining the learned features of a deep network is famously difficult, with abundant research to probe their inner workings (e.g.\citet{elhage2022toy, featurelearning-contrastive, tensor4}) and the relations between different models' learned representations \citet{pmlr-v97-kornblith19a, NEURIPS2018_a7a3d70c}. Similar to the studies in \citet{Fort2021ExploringTL}, \citet{uppaal-etal-2023-fine}, and \citet{jointfeatures}, we use PCA to perform dimensional reduction on data represented in high-dimensional space and extract key features to compare datasets and their underlying distributions.


\begin{section}{Background and Methods} \label{sec:methods}
\textbf{Data.} \label{sec:data-prep} All experiments utilize the Sentiment140 Dataset \citet{Go2009TwitterSC}, which is a collection of tweets scraped from Twitter in the year 2009 \footnote{\url{https://huggingface.co/datasets/sentiment140}}. The full dataset contains an automatically processed training dataset \(P\) and manually curated test set \(M\) which are collected using different methodologies. The 1,600,000 examples of $P$ have labels inferred from sentimentally relevant emoticons scrubbed from the original tweets, e.g. ``:)'' indicates a positive tweet and ``:('' negative. In contrast, \(M\) consists of 359 examples with human annotated sentiment labels.  There are two distinct sampling populations: \(P\) comes from the population of tweets containing emoticons, and \(M\) from a population of tweets containing references to a varied list of subjects generated by the dataset authors. This is problematic if the emoticons do not perfectly match onto the notion of language-sentiment; Appendix~\ref{app:table3} lists some datapoints from \(P\) that are likely confounders. The original authors behind Sentiment140 intended to view $P$ and $M$ as testing and training sets, respectively. As different experiments will train on samples coming from either $P$ or $M$, we refrain from referring to $P$ as `the' training set or $M$ as `the' testing set.



\textbf{Approximating Distribution Shift.} In this article we deal with neural networks trained on natural language which we use to predict binary sentiment classification. For simplicity, we present such a network as a function $F: \mathpzc{L}\to \{0,1\}$, where $\mathpzc{L}$ denotes the sampling space of natural language. The models considered here have a factorization $F = H \circ E$ where $E : \mathpzc{L}\to \mathbb{R}^d$ is a high-dimensional embedding obtained from the final hidden layer and $H : \mathbb{R}^d\to \{0,1\}$ is a classification head.

Given a sample of tweets $\{t_1, t_2, \ldots, t_n\}$ drawn from a distribution $p(t)$, we pass each tweet $t_i$ through $F$ and extract the feature activations $x_i := E(t_i) \in \mathbb{R}^d$, $i=1,\ldots,n$ following a distribution $q(x)$. The $x_i$'s serve as data engineered by the neural network with which we perform PCA to approximate $q(x)$. If we have another sample of tweets $\{t'_1,t'_2,\ldots,t'_{n'}\}$ drawn from $p'(t)$ then we can perform the same analysis to approximate the underlying distribution $q'(x)$ of the associated vectors $\{x'_1, x'_2,\ldots, x'_{n'}\}$. Importantly, we note that if $q(x) \ne q'(x)$, then we must have $p(t) \ne p'(t)$. That is, if the underlying distribution for the embedded samples is shifted, then so too must be the distribution for the original tweets. Further exposition is given in Appendix~\ref{sec:appendixmethods}.

\textbf{Training.}  \label{sec:training}
We use three different training methodologies in Section \ref{sec:results}. The different methodologies contrast whether we allow additional feature learning ($\FT$) or use the original pre-trained model's representations ($\LP$).
\begin{itemize}
    \item[$\FT)$] Starting with pre-trained weights, we attach a randomly initialized classification head and train for ten epochs, allowing \emph{all} weights to update. We use disjoint samples of $20,000$ points from $P$ for training and validation sets. The validation set is used to choose an early stopping epoch. Evaluation is performed on the entirety of $M$.
    \item[$\LP)$] Initialization as in $\FT$, allowing only the final classification layer to update. We use a balanced set of 300 points in $M$ for training the final layer, evaluating on the remaining 59.
    \item[$\FT+\LP)$] First train using the $\FT$ method from above, without evaluation. Then re-initialize the final layer, training and evaluating following the instructions in $\LP$.
\end{itemize}


\end{section}

\section{Results}\label{sec:results}
  
 
\subsection{LLMs can separate $P$ and $M$ with pre-trained weights.}\label{sec:batch} 
We use various popular open-source models \citet{Devlin2019BERTPO, Liu2019RoBERTaAR, falcon40b, llama2, Radford2019LanguageMA} to visualize the features of $P$ and $M$ using PCA on the final layer activations. We project down to the first two principal components and visualize the distribution with kernel density estimation. Results are shown in Figure \ref{PLOT1}. Five of the six language models are able to visually separate between $P$ and $M$. Both RoBERTa-base and Falcon 1B do so with some overlap of the embedded approximations while other models like GPT-2, Falcon 7B, and Llama-2 7B are able to highly differentiate the shift. This suggests that a model's ability to discriminate on distribution shifts may correlate with either model complexity, the training corpus, or an interaction thereof. 
  
  

  

\subsection{Fine-tuning on datasets similar to $M$ does not generalize to $M$.}
Here we compare additional feature learning $\FT$ to re-using the pre-trained model's representations $\LP$, by evaluating both methods effect on generalization on $M$. Table \ref{fig:ftGPT-2vsLP} indicates that the pre-trained features in GPT-2 have robust capabilities to classify on $M$. In Appendix~\ref{statistical} we undergo further analysis to give evidence that $\LP$ statistically outperforms $\FT$ ($p$-value 0.007). In context of Figure~\ref{PLOT1}, we view the relative performance statistics between $\LP$ and $\FT$ as a consequence of the misaligned features between $P$ and $M$. 

In addition to the performance gains from $\LP$ over $\FT$, we stress the substantial savings in both time and space complexity by simply training a linear classifier rather than fine-tuning all of the weights of model. Training via $\LP$ (using $M$) takes under 3.8 seconds/epoch with 2.4 GB VRAM compared to $\FT$ taking up to 11.9 minutes/epoch with 4.0 GB VRAM. 


\begin{table}[!ht]
\centering
\caption{Comparing Fine-Tuned GPT-2 with Targeted Linear Probes}
\begin{tabular}{lccc}
\toprule
Base Model & Method & Train Acc. (\%) & Test Acc. (\%) \\
\midrule
Pre-Trained GPT-2 & $\FT$ & $91.6_{\pm 0.93}$   & $84.1_{\pm 0.49}$\\
Random Features GPT-2 & $\LP$ & $85.7_{\pm 0.27}$  & $57.8_{\pm 1.05}$\\
Pre-Trained GPT-2  & $\LP$ & $95.3_{\pm 0.15}$ & $86.3_{\pm 0.55}$\\
Pre-Trained GPT-2  & $\FT + \LP$ & $93.9_{\pm 0.27}$    & $84.7_{\pm 1.17}$\\
\bottomrule
\end{tabular}
\label{fig:ftGPT-2vsLP}
\caption*{Table 1: Those equipped with the $\FT$ method are fine-tuned using 20k datapoints from $P$ and tested on all 359 points of $M$. Those with the $\LP$ method have a linear probe trained using 300 points of $M$ and tested on the remaining 59 points of $M$. Values are shown $\pm$ the standard error of the mean.}
\end{table}

\begin{section}{Discussion}


This article has presented a first approximation to visualizing the distributions of feature-embeddings, allowing us to confirm that a suspected distribution shift adversely affects a foundation model's representations. 
We used this tool to observe the shift between Sentiment140's automatically processed and manually curated subsets as represented by six foundation models. Five of the models' embeddings clearly separate in the first two PCs, despite a surface-level similarity in the natural text (Figure~\ref{PLOT1}). In fact, the separation between these distributions grows with model size, suggesting that greater model complexity allows better separability of text drawn from different sources. It is possible this effect is causally linked to other covariates between models, including training corpus and architectural improvements. Subsequent experiments with GPT-2 showed that features learned from fine-tuning on can harm predictive capacity on $M$ compared to a simple linear probe on $M$ which used the pre-trained features of the pre-trained model and only required 1.5\% of the data to train. This highlights both the potential consequences of training with misaligned data as well as the value of in-distribution data.  

Though we only focused on Sentiment140 here, the distinct sampling procedures behind $P$ and $M$ reflects a common situation in data science where it is costly or impossible to sample from the true distribution of interest \citet{lowresource1, lowresource2}. 
Towards addressing distribution shift for TL with foundation models, our work argues for two separate yet related pre-processing steps. First, ask whether the existing pre-trained representation recognizes the test data? If yes, then a simple linear probe may be sufficient. If no, we follow up by asking: is the train/test split identically distributed? If so, then cautiously proceed with FT. In the final negative case if not, re-evaluate the data generation methodology as otherwise, full fine-tuning on shifted data can confound interpretability for model generalization and downstream analysis.


\section{Limitations and Future Work}
A fundamental limitation of our approach is that it is presently an ad hoc qualitative measure to detect distribution shifts rather than a principled test statistic. As discussed above, our suggestion that the ability of foundation models to discern distribution shift is a result of the model complexity is a simple statement of correlation. It is also possible that the ability is derivative of the improvements/differences to pre-training corpora; which re-emphasizes the need for open-source publication of pre-training data. A deeper study of the model's architecture, its overall expressivity, and the data upon which it is trained may one day lead to quantitative hypothesis testing.

\section*{Acknowledgements}
The work of AD, MV, AE, and TC were partially supported by the Mathematics for Artificial Reasoning in Science (MARS) initiative via the Laboratory Directed Research and Development (LDRD) Program at PNNL. 

\end{section}
\clearpage

\bibliography{main}
\bibliographystyle{iclr2024_conference}

\newpage
\appendix
\begin{section}{Experimental Details}
 All experiments were conducted on an Nvidia DGX-2 on a single A100-40 GPU using Python 3.9, via PyTorch \citet{Paszke2019PyTorchAI}. Due to memory limitations, we limited batch size to 1, while non-standard for full fine-tuning, still attained robust results.

\begin{subsection}{Uncertainty Estimates} 
All experiments were run multiple times to ascertain both training and generalization variability. Seeds were manually set for the torch, numpy, and random modules. Experiments on \(M\) were performed 50x, with seeds 0-49, where \(M\) was shuffled with that specific seed, then split into training and evaluation sets. The experiment involving training on both $P$ and $M$ using the $\FT+\LP$ training method was performed 20x, with seeds 0-20. All other experiments were performed 10 times, with seeds 0-9, shuffling the data, then splitting accordingly. 
\end{subsection}

\begin{subsection}{Dataset and Model Sourcing}
We sourced our data and models from openly available sources such as Hugging Face and Papers With Code. We obtained all training and evaluation data used in this experiment from the Stanford Sentiment140 website\footnote{\url{https://huggingface.co/datasets/sentiment140}}. State-of-the-Art (SOTA) benchmarks were obtained from Papers With Code\footnote{https://paperswithcode.com/sota/text-classification-on-sentiment140}.  We used \textit{GPT-2ForSequenceClassification}, and \textit{GPT2Tokenizer} from Hugging Face for our experiments loaded from the Transformers Python library \citet{DBLP:journals/corr/abs-1910-03771}. 
\end{subsection}

\begin{subsection}{Pre-Processing Steps}
We pre-processed text in Sentiment140 according to steps outlined in \cite[Sec 2.3]{Go2009TwitterSC}. Duplicate tweets were removed. In addition, we used regular expressions to find usernames, remaining emoticons, and URLs.  For example, a given username in a tweet would be converted to the token USERNAME. Likewise, a url would be converted to the token URL. Remaining emoticons were removed during our pre-processing.  
\end{subsection}

\begin{subsection}{Feature Extraction for Distribution Shift} \label{sec:appendixmethods}
\textbf{LLMs and Feature Extraction.} In this article we deal with neural networks trained on natural language which we use to predict binary sentiment classification. For simplicity, we present such a network as a function $F: \mathpzc{L}\to \{0,1\}$, where $\mathpzc{L}$ denotes the sampling space of natural language. The models considered here actually have a factorization $F = H \circ E$ where $E : \mathpzc{L}\to \mathbb{R}^d$ is an embedding function into a high-dimensional euclidean space obtained from the final hidden layer and $H : \mathbb{R}^d\to \{0,1\}$ is a classification head obtained by applying a linear projection followed by softmax on the resulting logits.

Given a pre-trained model $F = H\circ E$ and a sample of tweets $\{t_1, t_2, \ldots, t_n\}$, we pass each tweet $t_i$ through $F$ and extract the feature activations $x_i := E(t_i) \in \mathbb{R}^d$, $i=1,\ldots,n$. The $x_i$'s serve as data engineered by the LLM which we then use for PCA; using the matrix $\Phi:=[x_1, x_2, \ldots, x_n]$, singular value decomposition gives a factorization $\Phi=U\Sigma V^{\intercal}$. The $k$-th principal component is then given by the $k$-th column of $V$.

\textbf{Distribution shift.} Let $X=\{(x_i, y_i)\}_{i\in I}$ (resp. $X'=\{(x'_i, y'_i)\}_{i\in I'})$ be a labeled training (resp. testing) dataset drawn from a distribution $p(x,y)$ (resp. $p'(x,y)$). We say that there is a distribution shift if $p(x,y)\ne p'(x,y)$. That is, $X$ and $X'$ are not drawn from the same distribution. Abusing language, in this case we also say there is a distribution shift between $X$ and $X'$. Fixing a choice of LLM and expressing it as $F=H\circ E$, we can study distribution shifts by working in the ambient embedding space, $\mathbb{R}^d$, by applying $E$. If the transformed distributions in this new space are distinct, then there must be a shift between the original distributions $p(x,y)$ and $p'(x,y)$. Examples of distribution shift appear throughout data science including covariate shift, sample-selection bias, and more \citet{10.5555/1462129}.

\end{subsection}

\begin{subsection}{Additional Experiments on Distributional Shift via Linear Classifiers} \label{sec:appendix bias linear probe}
In order to understand whether \(P\) and \(M\) are linearly separable, we create a dataset $(Bias)$ to identify the model's ability to classify the selection bias of Sentiment140. We choose 200 points from \(P\) and \(M\) respectively to form a training set \(B_{\text{train}}\) (balanced by sentiment labels in both categories), and we test generalization on 160 points of \(P\) and 159 points of \(M\), also balanced by sentiment labels, \(B_{\text{test}}\).\\

We perform linear probing for random feature, pre-trained, and fine-tuned initializations of GPT-2 training on the $(Bias)$ dataset. We train a logistic regression classifier with training data \(B_\text{train}\) for ten epochs. We use \(B_\text{train}\) to determine an epoch to evaluate test accuracy on \(B_{test}\).
\end{subsection}

\begin{subsection}{Additional Experiments on P and M}

We list our full set of experiments in table~\ref{tab:MainComparison}, some of which were not highlighted in the main text.
\newline
\newline
To understand the relative effect of training on $P$, we also test linear probes of random feature and pre-trained GPT-2 initializations. We take the same train/val/test split for sampling on $P$ as in \ref{sec:data-prep}, training on $P_{\text{train}}$, validating on $P_{\text{val}}$ and testing on $P_{\text{test}}=M$. 

We additionally investigate whether fine-tuning on $M$ would overfit compared to a linear probe. We fine-tune on $M$, taking the same train/test split as described for sampling on $M$ in \ref{sec:data-prep}, training on $M_{\text{train}}$ and testing on $M_{\text{test}}$. Furthermore, we perform linear probing with samples of 100-250 points of $M$, increment by 50 per iteration and test on a held-out set of 59 points. Throughout, we use the same hyperparameters as in our original experiments in \ref{sec:training}.
\end{subsection}

\begin{subsection}{Hyperparameters}
Hyperparameters used for all our experiments are shown in Table: \ref{tab:Hyperparameters}. \textit{BCEWithLogitsLoss} fuses a binary cross entropy loss and a sigmoid scaling operation. 
\begin{table}[!ht]
\centering
\caption{Hyperparameters used in linear probing and fine-tuning}
\begin{tabular}{llll}
\toprule
Hyperparameter & Linear Probing & Fine-Tuning \\
\midrule
Epochs & $10$ & $10$ \\
Optimizer  & Adam & Adam \\
Learning Rate & $10^{-2}$ & $10^{-5}$ \\
$\beta_1$ & $.9$ & $.9 $ \\  
$\beta_2$ & $.999$ & $.999$ \\ 
$\epsilon$ & $10^{-8}$ & $10^{-8}$ \\ 
Batch Size & $1$ & $1$ \\
Scheduler-Type & LinearLR & LinearLR \\ 
Start Factor & $.33$ & $.33$ \\
End Factor & $1.0$ & $1.0$ \\
Scheduler Iterations & $5$ & $5$ \\
Dropout & $0.1$ & $0.1$ \\ 
Gradient Norm Clip & $1.0$ & $1.0$ \\
Seeds & $0-49$ for $M$, $0-9 $ otherwise & $0-49$ for $M$, $0-9$ otherwise \\ 
Loss Function & BCEWithLogitsLoss & BCEWithLogitsLoss \\ 
Random Feature Initialization  & \makecell[l]{Sample from $\mathcal{N}(0,0.02)$,\\ residual layers scaled by $\frac{1}{\sqrt{n}}$} & -- \\
\bottomrule
\label{tab:Hyperparameters}
\end{tabular}
\end{table}

\end{subsection}
    \begin{subsection}{Visualizing the Methods FT and LP}
    Figures \ref{fig:ftGPTdiagram}, \ref{fig:LPdiagram}, and \ref{fig:RFdiagram} illustrate the experiments described in section \ref{sec:methods}. Specifically, figure \ref{fig:ftGPTdiagram} presents the typical transfer learning paradigm of learning on one dataset ($P$ in this case) and evaluating on a separate `expectedly similar' dataset ($M$). Figures \ref{fig:LPdiagram} and \ref{fig:RFdiagram} diagram attaching a linear head to examine feature representations for pre-trained and random feature initializations of GPT-2. 
        \begin{figure}
            \centering
            \includegraphics[width = .7 \linewidth]{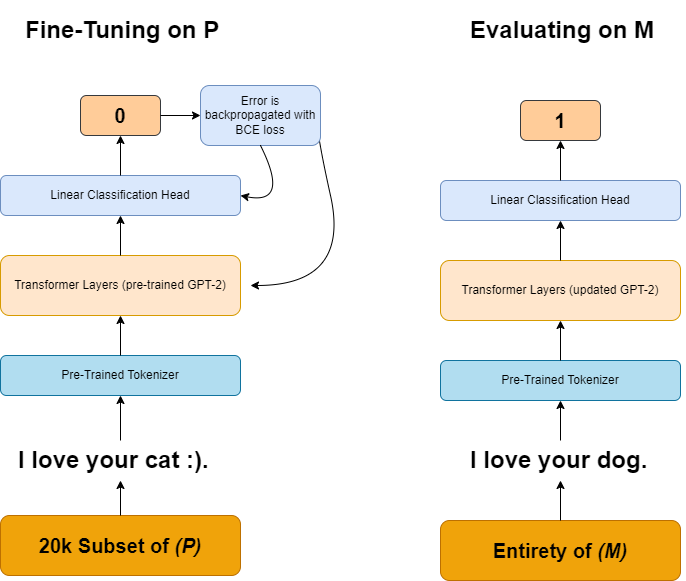}
            \caption{Full-fine-tuning of pre-trained GPT-2, trained on a sample of P, evaluated on M}
            \label{fig:ftGPTdiagram}
        \end{figure}
        \begin{figure}
            \centering
            \includegraphics[width = .7 \linewidth]{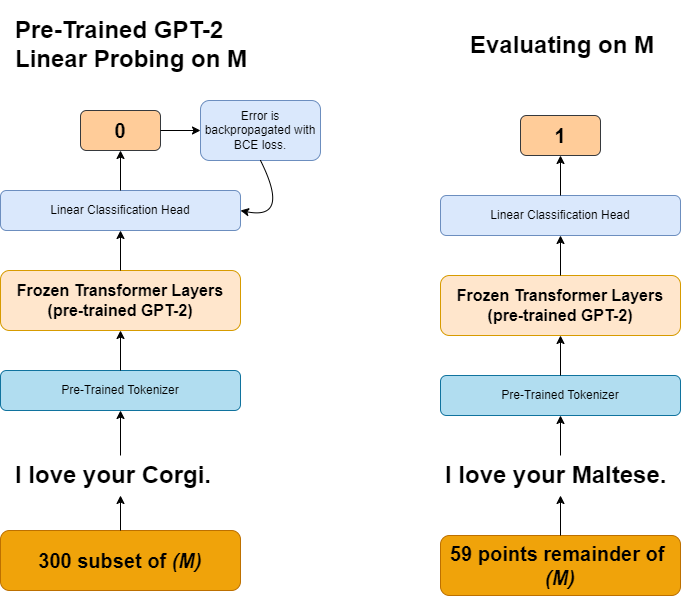}
            \caption{Linear probe of pre-trained GPT-2, trained on a sample of M, evaluated on remainder of M.}
            \label{fig:LPdiagram}
        \end{figure}
        \begin{figure}
            \centering
            \includegraphics[width = .7 \linewidth]{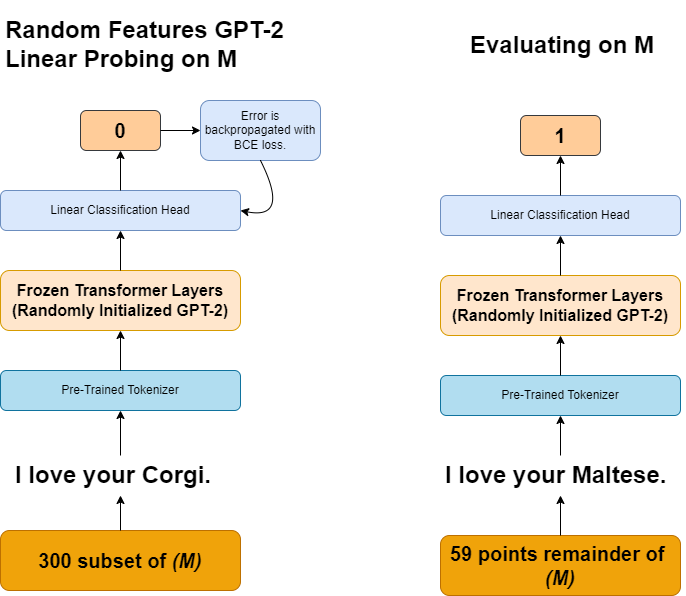}
            \caption{Linear Probe of Random Feature GPT-2, trained on a sample of M, evaluated on the remainder.}
            \label{fig:RFdiagram}
        \end{figure}
    \end{subsection}
\end{section}
\newpage

\begin{section}{Examining P for Examples of Mislabeled Data}\label{app:table3}
\begin{table}[!ht]
\centering
\caption{Possibly Mislabeled Data in $P$}
\begin{tabular}{lllll}
\toprule
Label & ID  & Tweet \\
\midrule
:( & 1467863072 & \makecell[l]{@twitterhandle1}  \\ 
:(& 1467871226 & \makecell[l]{@twitterhandle2 Congrats!! \\ i totally forgot to submit photos} \\ 
:(& 1467979094 & \makecell[l]{@twitterhandle3 OHH! OMG. LMAO. \\ I'm crying right now, LOL! KUTNERRRR was the best!}  \\ 
:) & 1836399555 & \makecell[l]{@twitterhandle4} \\ 
:) & 1879932456 & \makecell[l]{@twitterhandle5 ????????????????} \\ 
:) & 1833894264 & \makecell[l]{Just another SARS is coming...} \\ 
\bottomrule
\label{tab:Confounders}
\end{tabular}
\caption*{Table: 3 We present selected examples of potentially mislabeled in dataset $P$. The curators of Sentiment140 scrubbed the tweets in $P$ of emoticons. A ':(' corresponds to a 0 label, ':)' corresponds to 1 label.}
\end{table}    
In retrospect, using emoticons as a proxy to label sentiment is an imperfect method; one that the creators of Sentiment140 acknowledged as a key difficulty of creating their dataset \citet{Go2009TwitterSC}. Table \ref{tab:Confounders} showcases typical examples of such statements. Rows one and four consist of tweets only consisting of a social media handle after the relevant emoticon is removed. After pre-processing, the handle is changed to the token USERNAME, making these tweets a possible source of bias in the training data as inputs consisting of the exact same sequence of tokens would have opposite labels.
\newline
\newline
Row two includes an example of mixed sentiment. One aspect of the tweet is positive ("Congrats!!"), yet one is negative "i totally forgot to submit photos", nevertheless a human annotator may label this tweet as positive considering it starts by mentioning another user's handle and congratulating them. Row three is arguably an example of a false-negative, and we believe that this sentence expresses positive sentiment. To us, row six shows a false-positive, referencing the spread of a disease. In its original context, with a smiling emoticon, the tweet in row six would have been an example of sarcasm. Row five is also a potential false-positive as a long sequence of question marks is traditionally associated with disbelief or anger. Given that $M$ was hand-labeled, we can assume that false-negatives and false-positives as shown in Table \ref{tab:Confounders} would not be as prevalent in the data. 
\end{section}

\begin{section}{Highlighting a Random Feature Baseline}
  In their paper, Conneau et. al. \citet{Conneau2017SupervisedLO} demonstrate that linear probing of BiLSTM encoders initialized with random weights can achieve a peak of $80.7 \%$ accuracy in the SST-2 sentiment analysis dataset \citet{Socher2013RecursiveDM}. On our dataset, we average $57.8 \%$ test accuracy and reach $76\%$ peak as shown in Table \ref{fig:ftGPT-2vsLP} and Figure~\ref{fig:enter-RFD}. We extend the work of Conneau et. al. \citet{Conneau2017SupervisedLO} with LSTMs to LLM models such as GPT-2 in our case study.

 Recently, compelling work has been done investigating the generalization performance of random feature classification \citet{Mei2019TheGE}, \citet{Adlam2020UnderstandingDD}, \citet{Yehudai2019OnTP}. These are closely related to work on random initialization of neural networks as studied in \citet{Giryes2015DeepNN}. Daniely, Frostig and Singer \citet{Daniely2016TowardDU} show that with a random initialization, linear probing of the last hidden layer (equivalently last-layer training) can learn linear functions as well as more compliciated classes of functions obtained by non-linear kernel composition. These learning guarantees may be sufficient to achieve the generalization shown in our sentiment analysis task. In light of their work, studying random feature regression presents an exciting opportunity for AI research.
 \end{section}

\begin{section}{Additional Results}

\begin{table}[!ht]
\centering
\caption{Accuracy and Comparison to Other Works}
\begin{tabular}{llll}
\toprule
Model & Training Accuracy & Validation Accuracy & Test Accuracy \\
\midrule
\\
 RF GPT-2 + LP (\(P\))   & $60.5 \pm 0.20$\%   & $59.6 \pm 0.14$\% &   $53.6 \pm 0.62$\% \\
Pre-Trained GPT-2 + LP (\(P\))   & $77.3 \pm 0.1$\% & $76.4 \pm 0.09$\% &  $78.7 \pm 0.75$\% \\
 Pre-Trained GPT-2 + FT (\(P\)) & $91.6 \pm 0.93$ \% & $83.5 \pm 0.09$ \% & $84.1 \pm 0.49 $\%
 \\ \\
 \hline 
 \\
  RF GPT-2 + LP (\(M\)) &$85.7 \pm 0.27$\%  & --  & $57.8 \pm 1.05$\% \\
 Pre-Trained GPT-2 + LP (\(M\))& $95.3 \pm 0.15$\%  & --  & $86.3 \pm 0.55$ \% \\
 Pre-Trained GPT-2 + FT (\(M\)) &  $99.9 \pm 0.02$ \% & -- & $83.7 \pm 0.81$ \% \\
\\
 \hline
 \\
RF GPT-2 + LP (Bias) & $ 95.9  \pm 0.2$\%   & --   &$86.9 \pm 0.8$\% \\
Pre-Trained GPT2 + LP (Bias) &   $94.5 \pm 0.38$\%  & -- & $91.4 \pm 0.36$\% \\
FT GPT-2 + LP (Bias) & $93.4 \pm 0.28$ \% & -- & $92.3 \pm 0.64$ \% \\
 \\
 \hline
 \hline
 \\
  FT RoBERTa & -- & -- & $89.3$ \% \\
  FT ALBERT & -- & -- & $85.3$ \% \\
 FT XLNET & -- & -- & $84.0$ \% \\ 
\bottomrule
\label{tab:MainComparison}
\end{tabular}
\caption*{Table 4: Group one and two in the table are trained using data from $P$ and $M$, respectively. They are both evaluated using testing data from $M$. The third grouping is trained and tested on the Bias dataset described in \ref{sec:appendix bias linear probe}. The final group are state-of-the-art (SoTA) fine-tuned model instances described on the online repository Papers With Code.} 
\end{table}    

\begin{figure}
    \centering
    \includegraphics[width = .7 \linewidth]{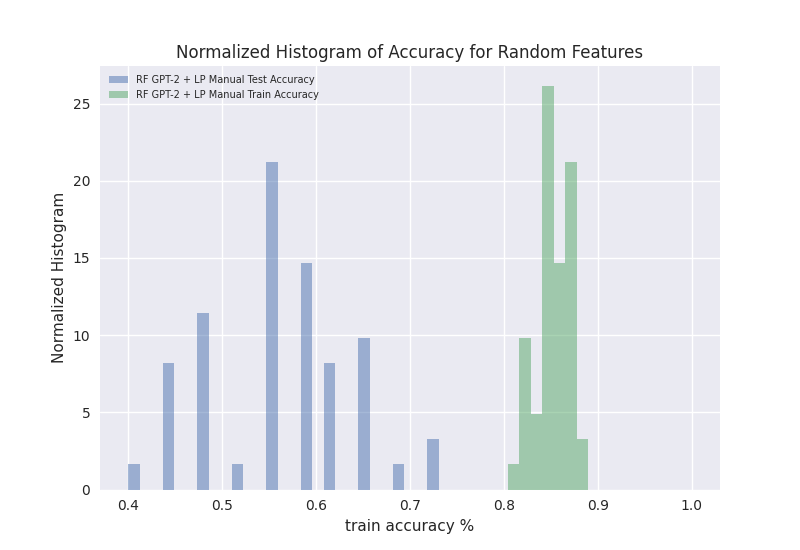}
    \caption{Training and Test Accuracy of a linear probe on the random features from GPT-2 architecture, trained on a held-out sample of 300 points from $M$, evaluated on the remainder.}
    \label{fig:enter-RFD}
\end{figure}
\begin{subsection}{Linear Probes separate P and M with high accuracy}
   \begin{table}[!ht]
\centering
\caption{Linear Probes of the $(Bias)$ dataset}
\begin{tabular}{llll}
\toprule
Base Model & Training Accuracy  & Test Accuracy \\
\midrule
 Random Features GPT-2 & $ 95.9  \pm 0.2$\%   &$86.9 \pm 0.8$\% \\
 Pre-Trained GPT-2 &   $94.5 \pm 0.38$\%  & $91.4 \pm 0.36$\% \\
 Fine-Tuned GPT-2 & $93.4 \pm 0.28$ \% & $92.3 \pm 0.64$ \% \\ 
\bottomrule
\label{tab:BiasComparison}
\end{tabular}
\caption*{We present linear probes from three different initializations trained to classify whether a tweet belonged to either $P$ or $M$. Our models were trained on $B_\text{train}$, and evaluated on $B_\text{test}$ from the $(Bias)$ dataset described in \ref{sec:appendix bias linear probe}.}
\end{table}   
Expanding on our exploratory analysis in \ref{sec:batch}, we trained a series of linear classifiers to determine if tweets belonged to either $P$ or $M$ according to the experimental setup described in \ref{sec:appendix bias linear probe}. Our results show that even a linear probe off of random features was able to perform this task with $86.9 \%$ accuracy as shown in \ref{tab:BiasComparison}. Moreover, we see that probes of the pre-trained and fine-tuned models can separate the distribution shift with even greater accuracy.
\end{subsection}
 
\begin{subsection}[Training on P under-performs on P compared to M]{Training on $\boldsymbol{P}$ under-performs on $\boldsymbol{P}$ compared to $\boldsymbol{M}$}

Using GPT-2's pretrained feature representations to examine $P$ and $M$ gives a surprising result. The validation accuracies on $P_\text{val}$ in rows 2 and 3 of Table \ref{tab:MainComparison} are lower than the testing accuracies on $M$. This result should give us pause by itself, since we typically expect the generalization metric to be lower than the training metric. This is especially the case with using $P$ and $M$ as training and testing distributions since our bias experiments indicate that $P$ and $M$ are linearly separable, hence are drawn from different distributions. We conjecture that the data of $M$ is more closely aligned to the pre-training corpus of GPT-2, than the data of $P$. Consequentially, one might view sentiment classification on $P$ as a more out-of-distribution task than sentiment classification on $M$. 

When these results are interpreted in context with the linear probe baseline of pre-trained GPT-2 on \(M\), we deduce that training on \(P\) may bias sentiment classification on \(M\) (also supported from the RF+LP on \(P\), row 1 of Table~\ref{tab:MainComparison}). Additionally, FT models on \(M\) show a clear overfit onto the training data (near $100\%$) with a substantial drop-off for test accuracy (approximately $83\%$).
\end{subsection}
\begin{subsection}{Training with more data on M leads to increased average performance}\label{statistical}
\begin{figure}
    \centering
    \includegraphics[width = .7 \linewidth]{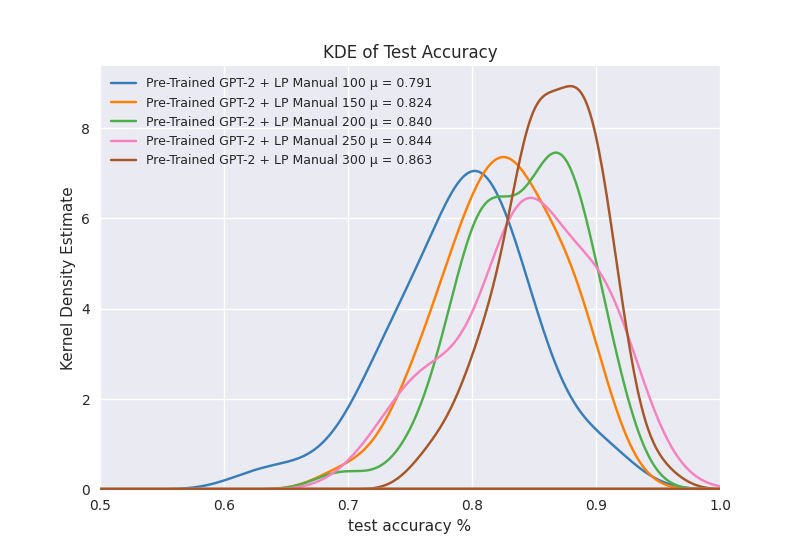}
    \caption{KDE plots of test accuracy for linear probes trained on 100/150/200/250/300 points of $M$ and tested on 59 points in the remainder of $M$}
    \label{fig:sample-size-lpm}
\end{figure}
Figure \ref{fig:sample-size-lpm} presents KDEs and means of the distributions of test accuracy for linear probing on varying size samples of $M$. As the sample size increases, we see a clear rightward trend. Moreover, going from 100 points to 300 points gives a relative gain of roughly $8 \%$. If this trend were to continue past 300 points, we could potentially see linear probe models with higher average test accuracy than existing state-of-the-art models.  
\end{subsection}
\begin{subsection}{FT on P and LP on M are not the same distribution}
\begin{figure}
    \centering
    \includegraphics[width = .7 \linewidth]{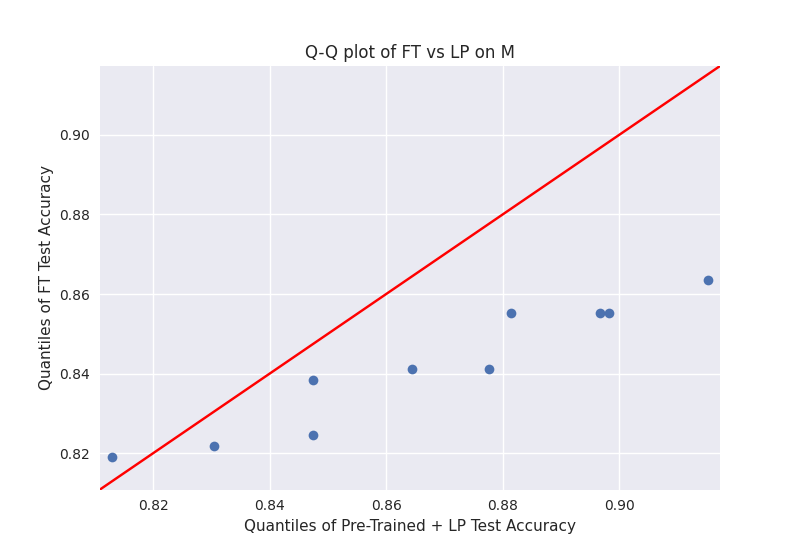}\\
    \includegraphics[width = .7 \linewidth]{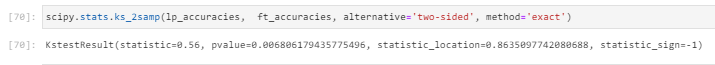}
    \caption{A QQ plot and a 2-sided KS test of test accuracies for linear probes of pre-trained GPT-2 and fine-tuned GPT-2 on M are presented.}
    \label{fig:distributiontests}
\end{figure}
The QQ plot for the quantiles of the test accuracies for LP on $M$ (on the x-axis) against FT on $P$ (on the y-axis) implies differing distributions with a larger dispersion for test accuracies of linear probes fit.  Moreover, applying a 2-sided KS test returns a \textit{p}-value of 0.007, showing that the distributions of fine-tuned models and linear probe models are likely not the same. 
\end{subsection}

\begin{subsection}{Features learned from $P$ do not improve the capacity to learn on $M$}
While there might be a slight skew on the QQ plot comparing $\FT(P)$ against $\FT(P)+\LP(M)$, there is no significant evidence of a statistical difference as seen by $p$-value on the KS-test. 
    \begin{figure}
    \centering
    \includegraphics[width = .4 \linewidth]{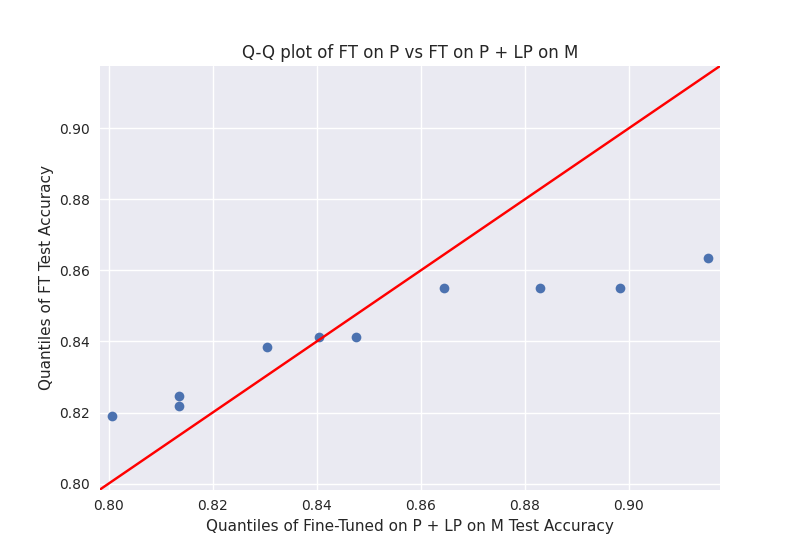}
    \includegraphics[width = .4 \linewidth]{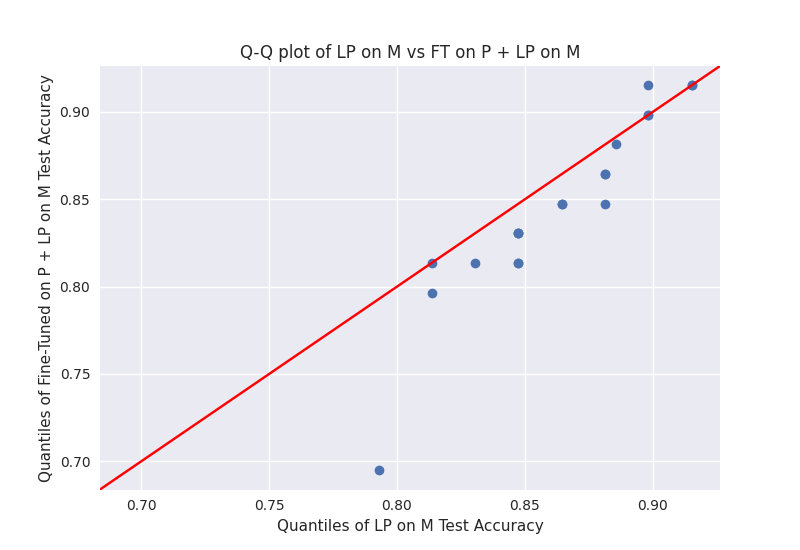}\\
    \includegraphics[width = 0.7 \linewidth]{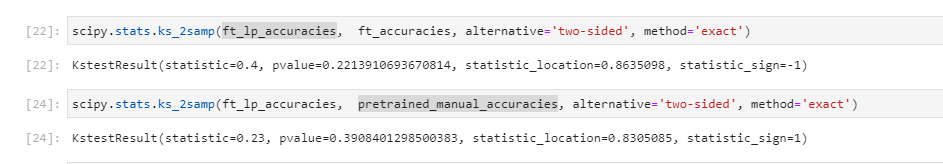}\\
    \caption{A QQ plot and a 2-sided KS test of test accuracies for linear probes of pre-trained GPT-2 and fine-tuned GPT-2 on M are presented.}
    \label{fig:distributiontests}
\end{figure}
\end{subsection}

\end{section}

\end{document}